\documentclass[letterpaper, 10 pt, conference]{ieeeconf}  

\IEEEoverridecommandlockouts                              

\usepackage{graphics} 
\graphicspath{ {./images/} }
\usepackage{epsfig} 
\usepackage{mathptmx} 
\usepackage{times} 
\usepackage{amsmath} 
\usepackage{amsfonts}
\usepackage{amssymb}  
\usepackage{wrapfig}
\usepackage{booktabs}
\usepackage{caption}
\usepackage{mathrsfs}
\usepackage{tabularx}
\usepackage{setspace}
\usepackage{subcaption}
\usepackage{graphicx}
\usepackage{color}

\title{\LARGE \bf
EffLoc: Lightweight Vision Transformer for Efficient 6-DOF Camera Relocalization
}

\author{ Zhendong Xiao$^{1}$, Changhao Chen$^{2}$, Shan Yang$^{1}$, $^*$Wu Wei$^{1}$ 
\thanks{$^{1}$Zhendong Xiao, Shan Yang and Wu Wei are with Control Science and Engineering,School of Automation Science and Engineering, South China University of Technology, Guangzhou, Guangdong Province, China
{\tt\small auxiao2022@mail.scut.edu.cn}}%
\thanks{$^{2}$Changhao Chen is with the Colloge of Intelligence Science and Technology, National University of Defense Technology, Changsha, China
{\tt\small changhao.chen66@outlook.com}}%
\thanks{*\textit{Corresponding author: Wu Wei}}
}

\begin{document}

\maketitle
\thispagestyle{empty}
\pagestyle{empty}

\begin{abstract}
Camera relocalization is pivotal in computer vision, with applications in AR, drones, robotics, and autonomous driving. It estimates 3D camera position and orientation (6-DoF) from images. Unlike traditional methods like SLAM, recent strides use deep learning for direct end-to-end pose estimation.
We propose EffLoc, a novel efficient Vision Transformer for single-image camera relocalization. EffLoc's hierarchical layout, memory-bound self-attention, and feed-forward layers boost memory efficiency and inter-channel communication. Our introduced sequential group attention (SGA) module enhances computational efficiency by diversifying input features, reducing redundancy, and expanding model capacity. EffLoc excels in efficiency and accuracy, outperforming prior methods, such as AtLoc and MapNet. It thrives on large-scale outdoor car-driving scenario, ensuring simplicity, end-to-end trainability, and eliminating handcrafted loss functions. 
\end{abstract}

\section{INTRODUCTION}

Camera relocalization (i.e. camera pose regression) focuses on the retrieval of the 3D position and orientation (6-DoF) of a camera based on the input images. It plays a crucial role in intelligent systems, ranging from augmented reality (AR) [1]/mixed reality (MR), delivery drones, robotics to autonomous driving [2]. 
Camera localization approaches have historically leaned on image structure and feature to match visual observation against a map [3], which establishes dense correspondences between 2D pixels and 3D points within the scene. Subsequently, the camera pose is estimated by employing  Perspective-n-Point (PnP) solver  [4] or the Kabsch algorithm with RANSAC [5]. These conventional relocalization methods fundamentally rely on a matching procedure, encompassing the comparison of a query image against a database of reference images [6]. The computational and storage requirements of these techniques correlate directly with the volume of sample points within the database. Furthermore, the efficacy of these methods is inherently intertwined with the quality of the matching process predicated upon the similarity score.

\begin{figure}[t]
    \centering
    \includegraphics[width=\columnwidth,height=0.6\columnwidth]{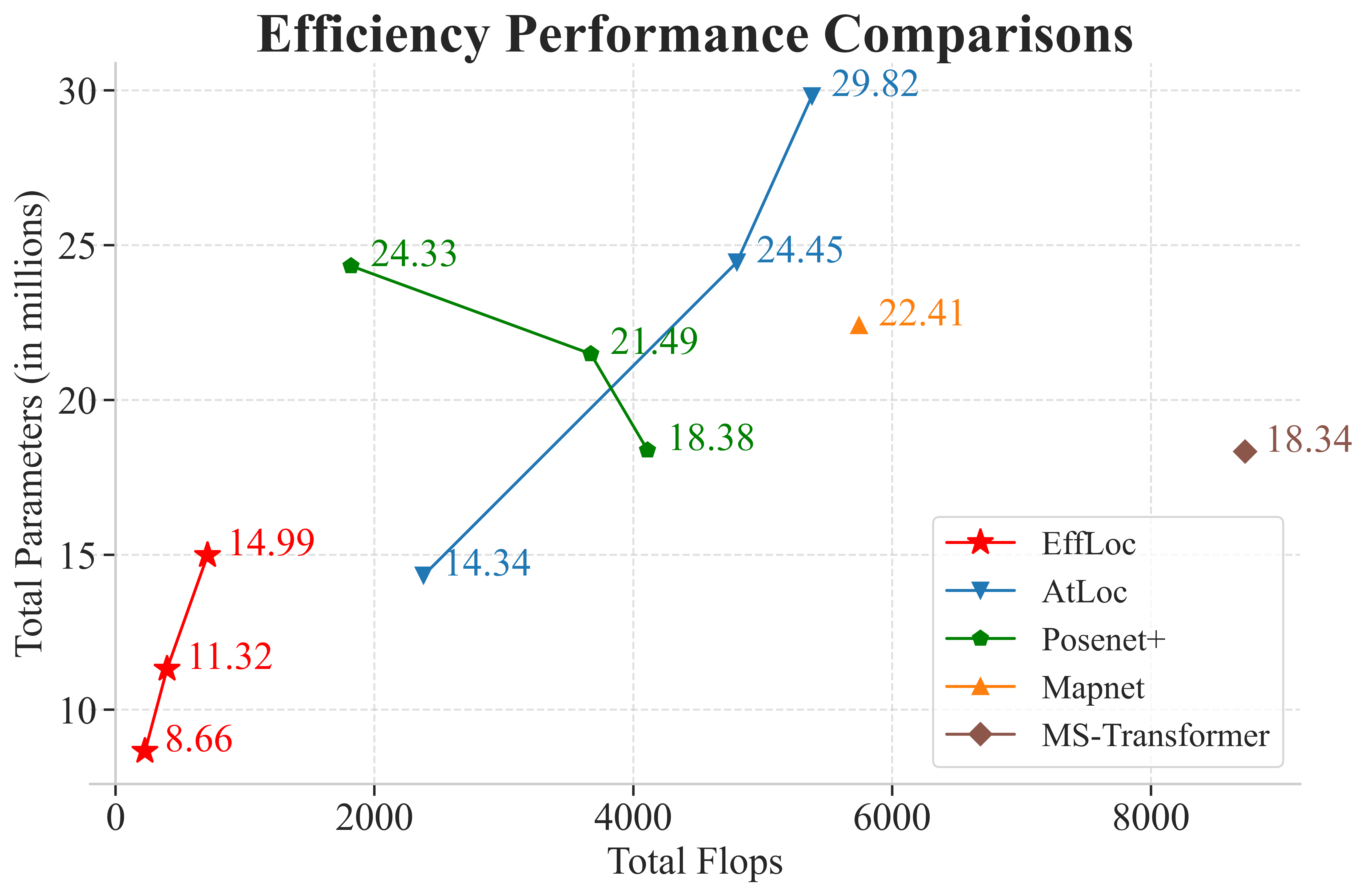}
    \captionsetup{font={footnotesize},justification=justified}
    
    \caption{A comparison between our proposed EffLoc model and other deep learning based relocalization models. The x-axis denotes Total Floating-Point Operations Per Second (FLOPs), while the y-axis represents the total number of parameters. The parameter count is labeled for each corresponding point on the graph. Our EffLoc models exhibit superior efficiency and computational complexity, attaining the lowest Flops and parameter counts.}
    \label{fig1}
\end{figure}


Deep learning-based camera relocalization methods can achieve end-to-end pose estimation directly from images via deep neural networks. For example, PoseNet [7] uses a convolutional neural network (CNN) based encoder to extract features from a single image as vector embeddings, which are subsequently transformed to 6-DoF pose. Other end-to-end leaning approaches such as [8] utilize an implicit map database to store scene information and eliminate the complex handcrafted feature engineering. 
Traditionally, deep learning based pose estimation heavily relies on Convolutional Neural Networks (CNNs) for feature extraction, which operate within localized pixel neighborhoods. However, Vision Transformers (ViTs) represent a recent breakthrough by segmenting images into patches and utilizing position embeddings to capture global dependencies. Unlike CNNs, ViTs establish meaningful correlations among spatially distant image regions, crucial for real-world relocalization tasks with large datasets. In addition, CNN-based visual localization models suffer from accuracy limitations and lack robust generalization due to challenges like lighting variations, occlusions, and dynamic objects. In contrast, ViTs, when trained with image-poses pairs, align better with map libraries, eliminating scale drift and cumulative errors. The emerging Light-weight Vision Transformers [10] offer computational efficiency and improved robustness in complex, resource-constrained real-world scenarios.

\begin{figure*}[t]
    \centering
    \includegraphics[width=\textwidth,height=0.37\textheight]{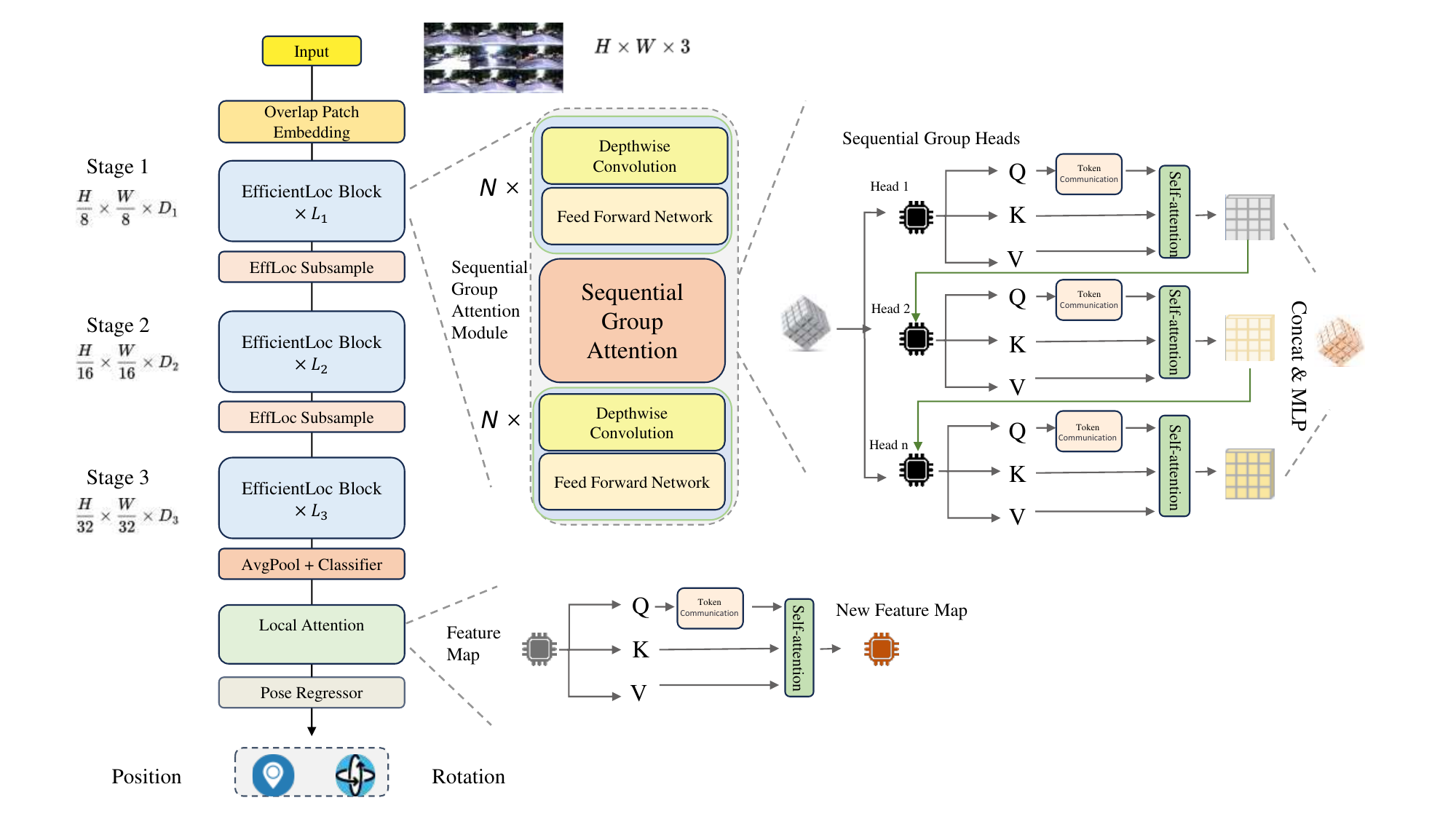}
    \captionsetup{font={footnotesize},justification=justified}
    \caption{An overview of EffLoc's hierarchical framework and its modules. The left column showcases the overall layout. The middle column highlights Sequential Group Attention Module and Sequential Group Heads (SGH). The right column details how SGH integrates outputs across heads. The bottom presents attention feature map and the pose regressor overview for feature-to-pose transformation.}
    \label{fig2}
\end{figure*}


In addressing the challenges posed by existing CNN-based visual relocalization methods, we introduce EffLoc, a lightweight Vision Transformer framework for efficient 6-DoF camera relocalization. Figure 2 presents a modular overview of the EffLoc. Our hierarchical architecture integrates memory-bound self-attention and inter-channel communication for improved memory efficiency [12] [13]. We also propose the sequential group attention (SGA) module, enhancing computational efficiency by diversifying input features across attention heads and redistributing parameters. Our model surpasses previous techniques, excelling in accuracy and efficiency trade-off for image-based 6-DoF camera relocalization, even on large-scale outdoor datasets. Its simplicity, end-to-end trainability, and elimination of hand-crafted loss functions mark its strengths.
Figure \ref{fig1} demonstrates the superior performance of our model in contrast to prior techniques, excelling in the trade-off between accuracy and efficiency in image-based 6-DoF camera relocalization. Notably adept at processing extensive outdoor datasets, our model shines with its streamlined architecture, end-to-end trainability, and elimination of the necessity for manually engineered geometric loss functions.

In summary, the contributions of this work are three-fold:

\begin{itemize}
\item We propose EffLoc, a novel light-weight end-to-end Vision Transformer architecture  6-DoF camera relocalization using single images that can generalize to large-scale real-word environments. 
\item We present a simple and effective sequential group module that remarkably enhances the latency/accuracy trade-off for image-based 6-DoF camera relocalization. By introducing diverse channel-wise feature splits across attention heads, this module effectively reduces redundant attention computations, leading to notable memory efficiency gains.
\item We introduce a  novel parameterization approach that involves reconfiguring the original Vit QKV (Query, Key, Value) ratios for camera pose estimation. This optimization substantially enhances computation and memory efficiency, resulting in a significant reduction in both FLOPs (86.8\%) and memory usage (49.7\%) compared to AtLoc [16]. Notably, the impact of Q and K in the third block is substantially diminished through this reconfiguration process.
\end{itemize}

\section{RELATED WORK}
\subsection{Deep learning for Camera Relocalization}
Deep learning has revolutionized camera relocalization, employing Convolutional Neural Networks (CNNs) to manage variations in illumination, viewpoint, and object occlusion [17], [18]. Kendall [19] first demonstrated end-to-end pose estimation using CNNs, eliminating intermediate steps like feature matching [20]. Furthermore, Deep Neural Network (DNN)-based camera localization methods obviate the need for manual construction of a map or a database of landmark features [21]. This approach evolved to include Recurrent Neural Networks (RNNs) and Bayesian CNNs for spatial-temporal accuracy and uncertainty estimation [22], [23]. 
CMRNet [46] and CMRNet++ [47], integrate deep learning with geometric approaches to taskle image-based relocalization within LiDAR maps.
However, challenges with hand-tuned scale factors and obstacles led to innovations like PosenetV2's geometric reprojection loss [20] and BranchNet's locator with two branches for position and angular deviations[24]. Attention mechanisms in Atloc [16] and additional reconstruction branches in MMLNet [25] further enhanced the relationship between 2D images and 3D scenes. Camera Pose Auto Encoders (PAEs) introduced lightweight test-time optimization [26]. Our EffLoc model, with its hierarchical layout of self-attention and feedforward layers, outperforms previous approaches with less computation costs, faster convergence velocity and achieves the best results in common benchmarks.

\subsection{Vision Transformers}
Transformers, introduced by Vaswani [11], have excelled in NLP and Computer Vision, surpassing RNNs in NLP [27]. Vision Transformers (ViTs), introduced by Touvron [28], segment images into patches with position encoding, achieving high performance on datasets like JFT-300M [29]. However, original ViTs faced optimization, efficiency, and training challenges [27] [28]. Lightweight transformers prioritize computational efficiency for real-time applications and constrained devices [14]. MobileNet [32] uses depthwise separable convolutions, TinyBERT [33] distills compact BERT [34] for constrained deployment, MobileViT V1 [35] explores channel, spatial factorizations, hierarchical token-to-token attention [36], adaptive token mixing [37]. ParC-Net [38] introduces patch-aware adaptive receptive fields, and EfficientVIT [39] innovates cascaded group attention. Next-ViT [10] addresses sparse attention, redundancy, complexity, memory. Our work integrates lightweight vision transformers, efficient parameterization into camera relocalization, demonstrating hierarchical group attention's robust key feature correlation, offering deployment-friendly solutions for accurate, high-speed camera relocalization. 

\begin{figure*}[t]
    \centering
    \includegraphics[width=2\columnwidth,height=\columnwidth]{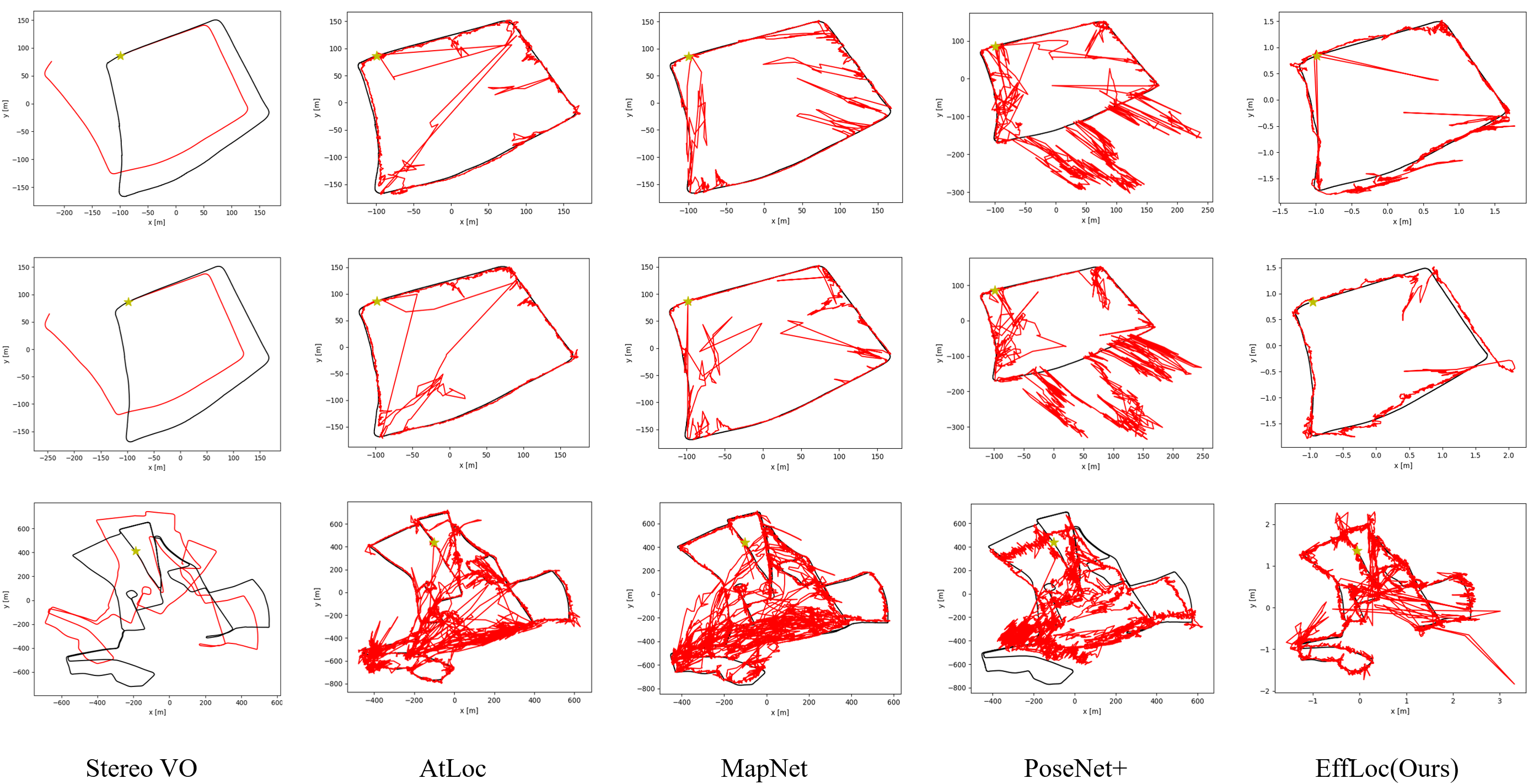}
    \captionsetup{font={footnotesize},justification=justified}
    \caption{Trajectories on LOOP1 (top), LOOP2 (middle), and FULL1 (bottom) of Oxford RobotCar. The black lines depict the ground truth trajectories, and the red lines represent the trajectory predictions. A yellow star denotes the starting point in each trajectory.}
    \label{fig3}
\end{figure*}

\section{LIGHTWEIGHT-TRANSFORMER BASED EFFICIENT CAMERA RELOCALIZATION}

This section presents an \textbf{Eff}icient Lightweight-Transformer based Camera Re\textbf{Loc}alization (EffLoc) approach, 
to learn 6-DoF camera poses from single images.

\subsection{Overlap Patch Embedding for Feature Extraction}
Vision Transformers (ViTs) are designed to capture holistic contextual information across images, distinct from Convolutional Neural Networks (ConvNets) that emphasize local features. Effective pose regression relies on extracting features from single images. In standard ViTs, input images are split into non-overlapping 16×16 patches, linearly projected into the encoder's input dimension using a learned weight matrix [27].
Our approach involves dividing images into overlapping patches, embedding each using a conventional ConvNet with layered convolutions [40]. This Overlap Patch Embedding enhances fine-grained localization by capturing local details and spatial sensitivity from neighboring patches.
Past studies [17] [38] demonstrate classical convolutional networks' effectiveness. [16] underscores ResNet34's efficacy, a 34-layer residual network, as a base for camera pose estimation. Residual networks like ResNet34 train deeper layers, addressing gradient vanishing and memory inefficiency issues associated with cross-memory access, which is computationally expensive. Hence, we opt for a lightweight Vision Transformer (EfficientVit) as EffLoc's backbone.
EfficientVit's weights are initialized from a pretrained ImageNet-1K model [42], optimized for image classification. Given an image \(I \in R^{C \times H \times W}\), the features \(X \in R^C\) can be extracted via the overlap patch embedding (Ope): 
$$
X = \mathrm{Ope}(I).\eqno{(1)}  
$$

\subsection{Hierarchical Layout to Enhance Memory Efficiency}
Here, we introduce hierarchical layout with competitive convergence capability that enhances memory efficiency and mitigates attention computation redundancy. Specifically, it applies a single less memory-bound self-attention layer \(\mathcal{L}_i^{SAL}\) with linear complexity and a depthwise convolution layer for spatial integration, which is nested between the feed forward layer \(\mathcal{L}_i^{FFL}\). The computation can be expressed as follows:
\[
X_{i+1} = \prod^{\mathit{N}}{\mathcal{L}_i^{FFL}\left(\mathcal{L}_i^{SAL}\left(\prod^{\mathit{N}}{\mathcal{L}_i^{FFL}\left(X_i\right)}\right)\right)},  \eqno{ (2)}
\]
where $X_i$ represents the complete input feature for the $i-th$ position. The hierarchical layout converts $X_i$ into $X_{i+1}$ with $\mathit{N}$ feed forward layers, both prior to and subsequent to the self-attention layer. This layout reduces the memory footprint while preserving crucial information and optimizes the utilization of model parameters resulting from self-attention layers. {By facilitating more efficient multiple feed forward network layers communication between different feature channels, the model improve the velocity of convergence, and thus reduce the computational burden.}

\subsection{Sequential Group Attention to Aggregate Features}
The redundancy of attention heads in multi-head self-attention is a significant issue that leads to computational inefficiency [15]. We introduce Sequential Group Attention (SGA) as attention module into our proposed EffLoc. Each head receives different subsets of the complete features, thereby effectively decomposing the attention computation across multiple heads in the attention module. Mathematically, this attention mechanism can be expressed as:

$$
\widetilde{X}_{ij} = \text{Attn}\left(X_{ij}\omega_{ij}^\mathrm{Q}, X_{ij}\omega_{ij}^\mathrm{K}, X_{ij}\omega_{ij}^\mathrm{V}\right),  \eqno{ (3)}
$$

where the $j$-th attention head performs self-attention computation over $X_{ij}$. The projection layers $\omega_{ij}^\mathrm{Q}$, $\omega_{ij}^\mathrm{K}$, and $\omega_{ij}^\mathrm{V}$ transform the input feature split into distinct subspaces. Finally, the self-attention of input features $X$ can be written as:

         $$ {\widetilde{X}}_{i+1}=Softmax\left(Concat\left[{\widetilde{X}}_{ij}\right]_{j=1:n}\right)\omega_i^\mathrm{L},                  \eqno{ (4)}$$

where $n$ is the total number of attention heads, i.e., $X_{i} = \left[X_{i1}, X_{i2}, \ldots, X_{in}\right]$ and $1 \leq j \leq n$. The linear layer $\omega_i^\mathrm{L}$ projects the concatenated output features back to the dimension consistent with the input, {ensuring dimensional aggregation coherence.} The SoftMax function is applied to normalize the attention scores and convert them into a probability distribution within the range $[0,1]$.

\subsection{Sequential Group Heads to Improve Feature Representation}
While utilizing feature splits instead of the complete features for each attention head enhances efficiency and reduces computational overhead, we aim to further enhance its capacity by encouraging the Q, K, and V layers to learn projections on feature representations that contain more comprehensive information. Figure 2 showcases sequential group heads (SGH) in to calculate the attention map which combine the output of each attention head with the subsequent head to iteratively refine the feature representations:
$$
SGH (X_{ij}) ={\ X}_{ij}+{\widetilde{X}}_{i\left(j-1\right)},1<j\le n, \eqno{ (5)}
$$
where $\text{SGH}(X_{ij})$ is the summation of the $\left(j-1\right)$-th head output ${\widetilde{X}}_{i\left(j-1\right)}$ and the $j$-th position input $X_{ij}$. 
Sequential Group Heads enhances the feature representations by adding the output of each head to the subsequent head. {An extra token interaction layer, utilizing depth-wise convolution, is incorporated before each feed-forward layer. This approach implements inductive bias to enhance the representation of local and global features}.

\subsection{Learning Camera Pose}
Our work is built upon prior works in VidLoc [43] pose estimation method, which regresses 6-DoF camera pose from Sequential Group heads guided features $\text{SGH}\left(X_{ij}\right)$ through Multilayer Perceptrons (MLPs):
$$\left[p,q\right]=MLPs\left(SGH\left(X_{ij}\right)\right).\eqno{ (6)}$$

Here $p$ represented by the 3D camera position and a 4D unit quaternion $q$ for orientation. The parameters inside the sequential group attention modules are optimized with L1 Loss via the following loss function [41]:
$$
loss\left(I\right)=|p-\hat{p}|_1e^{-\alpha}+\alpha+|\log{q}-\log{\hat{q}}|_1e^{-\beta}+\beta .\eqno{ (7)}$$

Here, $\alpha$ and $\beta$ balance position and rotation losses. The logarithm of a unit quaternion, denoted as $\log{q}$, offers a three-dimensional, minimally parameterized representation. This characteristic enables direct utilization of L1 distance as the loss function without normalization. The L1 loss reduces outlier impact, enhancing robustness to atypical observations and promoting parameter and feature sparsity. This encourages feature selection and the allocation of zero weights to irrelevant or non-significant features.

Specifically, the unit quaternion $q = (u, v)$ is represented with a scalar $u$ for the real part and a three-dimensional vector $v$ for the imaginary part, defined as:

\begin{align}
\log \mathbf{q} = \begin{cases}
    \frac{\mathbf{v}}{\|\mathbf{v}\|} \cos^{-1} u, & \text{if } \|\mathbf{v}\| \neq 0 \\
    \mathbf{0}, & \text{otherwise} 
\end{cases}
\tag{8}
\end{align}  

Quaternions are commonly used for camera pose regression due to their continuous and differentiable orientation representation. By normalizing 4D quaternions to unit length, any 3D rotation can be mapped to valid unit quaternions. However, quaternion non-uniqueness arises: $-q$ and $q$ can represent the same rotation due to the two hemispheres. To ensure uniqueness, this study constrains quaternions to a single hemisphere.

\begin{table*}[t]
  \centering
  \footnotesize
    \begin{tabular}{c|cc|cc|cc|cc}
    Dataset & \multicolumn{2}{c}{PoseNet+} & \multicolumn{2}{c}{MapNet} & \multicolumn{2}{c}{AtLoc} & \multicolumn{2}{c}{\textbf{EffLoc (Ours)}} \\
    \hline
    \hline
    —     & Median & Mean  & Median & Mean  & Median & Mean  & Median & Mean \\
    LOOP1 & 6.91m, 2.06° & 25.39m, 17.49° & 5.79m, 1.54° & 8.79m, \textbf{3.53°} & 5.68m, 2.23° & 8.61m,4.58° & \textbf{5.23m, 2.18°} & \textbf{7.58m,} 3.72° \\
    LOOP2 & 5.83m, 2.05° & 28.89m, 19.65° & 4.93m, 1.67° & 9.81m, 3.86° & 5.05m, 2.01° & 8.86m, 4.67°, & \textbf{4.76m,} 2.06° & \textbf{7.89m, 4.19°} \\
    FULL1 & 107.8m, 23.5° & 125.6m, 28.10° & 17.91m, 6.68° & 41.2m, 13.5° & 11.1m, 5.28° & 29.6m, 12.4° & \textbf{10.28m, 4.98°} & \textbf{27.23m, 11.41°} \\
    FULL2 & 101.9m, 21.1° & 131.1m, 26.55° & 20.34m, 6.39° & 59.5m, 14.7° & 12.2m, 4.63° & 48.2m, 11.1° & \textbf{11.12m, 4.17°} & \textbf{44.82m, 9.87°} \\
    \hline
    Average & 55.61m, 12.2° & 77.75m, 22.95° & 12.24m, 4.07° & 29.8m, 8.79° & 8.54m, 3.54° & 23.8m, 8.19° & \textbf{7.85m, 3.35°} & \textbf{21.88m, 7.40°} \\
    \end{tabular}%
    \captionsetup{font={footnotesize},justification=justified}
    \caption{Camera Relocalization results of the LOOP and FULL trajectories on the Oxford Robotcar dataset. Median and mean errors of position and rotation are calculated for each trajectory using Posenet+, MapNet, AtLoc, and our proposed EffLoc.}
  \label{tab1}%
\end{table*}%

\section{EXPERIMENTS}
\subsection{Experiment Setup}
\subsubsection{Implementation Details}
For consistent network training, images are rescaled via cropping to 256 × 256 pixels using random and central strategies. Input images are then normalized within a -1 to 1 intensity range. Pretrained model experiments rely on ImageNet-1K [42] data.
Model construction employs PyTorch 1.11.0 and Timm 0.6.13, training from scratch for 340 epochs on an Nvidia V100 GPU. We use AdamW [44] optimizer with a cosine learning rate scheduler. During Oxford Robot-car dataset training, we apply the ColorJitter augmentation technique, adjusting brightness, contrast, saturation, and hue (0.7, 0.7, 0.7, 0.5).
This augmentation enhances the model's ability to generalize across weather and time variations, making EfficientLoc robust across real-world scenarios (Figure 4).
The initial learning rate is $1 \times 10^{-3}$ with the following hyperparameters: weight decay of $3.5 \times 10^{-2}$, mini-batch size of 64, dropout rate of 0.5, and weight initializations of $\alpha = -5.0$ and $\beta = -1.0$.

\subsubsection{Oxford RobotCar Datasets and Baselines}
The Oxford Robot-Car dataset [45] offers a wealth of data with over 100 iterations of a 10km consistent route in central Oxford. This extensive dataset was captured biweekly for more than a year, encompassing a diverse array of environmental conditions, including weather variations, traffic dynamics, pedestrians, construction activities, and roadwork scenarios. Comprising images from six car-mounted cameras, the dataset integrates LIDAR, GPS, INS measurements, and stereo visual odometry (VO). Notably, the dataset presents dynamic elements, including mobile and stationary vehicles, cyclists, and pedestrians, thereby posing significant challenges for vision-based relocalization tasks.

To ensure an equitable comparison, we adopt the evaluation strategy established by MapNet [41]. Our experiments focus on two distinct subsets from this dataset: LOOP (1120m) and FULL (9562m), segregated based on route lengths. For comprehensive information regarding these sequences, consult Table \ref{tab2}.

\begin{table}\footnotesize%
  \centering
  \setstretch{1.1} 
    \begin{tabularx}{\columnwidth}{c|cccc}
    \toprule
    Sequence & Time  & \hspace{0.3em}Tag
    & \hspace{0.3em}Distance   & \hspace{0.3em}Mode \\
  \hline
    –     & 2014/06/26 8:53 & \hspace{0.3em}Cloudy & \hspace{0.3em}1120m & \hspace{0.3em}Training \\
    –     & 2014/06/26 9:24 & \hspace{0.3em}Cloudy & \hspace{0.3em}1120m & \hspace{0.3em}Training \\
    LOOP1 & 2014/06/23 15:41 & \hspace{0.3em}Sunny & \hspace{0.3em}1120m & \hspace{0.3em}Testing \\
    LOOP2 & 2014/06/23 15:36 & \hspace{0.3em}Sunny & \hspace{0.3em}1120m & \hspace{0.3em}Testing \\
    \hline
    –     & 2014/11/28 12:07 & \hspace{0.3em}Cloudy & \hspace{0.3em}9562m & \hspace{0.3em}Training \\
    –     & 2014/12/02 15:30 & \hspace{0.3em}Cloudy & \hspace{0.3em}9562m& \hspace{0.3em}Training \\
    FULL1 & 2014/12/09 13:21 & \hspace{0.3em}Cloudy & \hspace{0.3em}9562m& \hspace{0.3em}Testing \\
    FULL2 & 2014/12/12 10:45 & \hspace{0.3em}Cloudy & \hspace{0.3em}9562m& \hspace{0.3em}Testing \\
    \end{tabularx}%
    \captionsetup{font={footnotesize},justification=justified}
    \caption{Training and testing datasets from the Oxford Robot-Car dataset. The testing datasets in LOOP sequence recorded under direct sunlight, whereas FULL datasets are captured under cloudy conditions.}
  \label{tab2}%
\end{table}%

\begin{figure}[t]
  \centering
   \begin{subfigure}[b]{0.45\columnwidth}
    \centering
    \includegraphics[width=1\columnwidth,height=0.7\columnwidth]{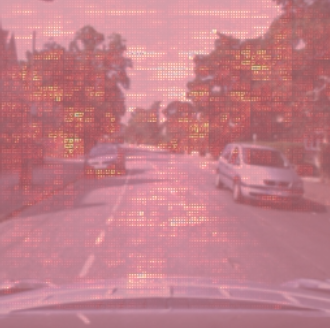}
    \label{fig4-1}
  \end{subfigure}%
  \hspace{0.02\columnwidth}
  \begin{subfigure}[b]{0.45\columnwidth}
    \centering
\includegraphics[width=1\columnwidth,height=0.7\columnwidth]{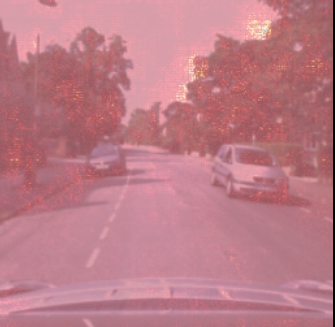}
    \label{fig4-2}
  \end{subfigure}
  \begin{subfigure}[b]{0.45\columnwidth}
    \centering
\includegraphics[width=\columnwidth,height=0.7\columnwidth]{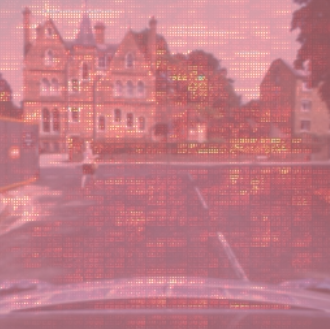}
    \subcaption{AtLoc}
    \label{fig7}
  \end{subfigure}%
  \hspace{0.02\columnwidth}
  \begin{subfigure}[b]{0.45\columnwidth}
    \centering
\includegraphics[width=1\columnwidth,height=0.7\columnwidth]{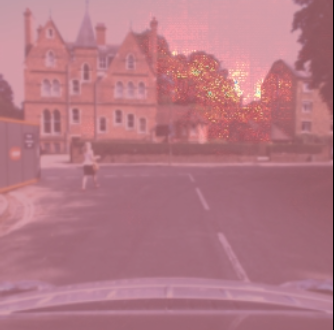}
    \subcaption{EffLoc(Ours)}
    \label{fig8}
  \end{subfigure}
   \captionsetup{font={footnotesize},justification=justified}
    \caption{Saliency maps of a representative scene from the Oxford RobotCar dataset illustrate EffLoc's adeptness in guiding the attention of the lightweight transformer towards geometrically resilient objects (e.g., distant skyline and trees edges on the right). This contrasts with environmental dynamics (e.g., the road in the top figure and moving pedestrians in the bottom), as observed in comparison with AtLoc. This emphasis contributes to enhanced global localization robustness.}
\end{figure}

\subsection{Visual Reloclization in Car-Driving Scenario}
The Oxford Robot-car dataset presents significant challenges due to its prolonged data collection period, necessitating a relocalization model with high robustness and adaptability. Table \ref{tab1} displays a comparison of our proposed methods against PoseNet+, MapNet, and Atloc. Compared with PoseNet+, EffLoc demonstrates remarkable improvements on both LOOP sequences and FULL sequences. Additionally, we conducted experiments involving Ms-transformer [9] on the Oxford Robot-car dataset. However, their model lacks effective generalization in large-scale testing scenarios, leading to ineffective results.The mean position accuracy is enhanced from 25.39m to 7.58m on LOOP1 and from 28.89m to 7.89m on LOOP2. EffLoc reduces the mean rotation error from 17.49° to 3.72° on LOOP1 and from 19.65° to 4.19° on LOOP2. EffLoc achieves the largest performance gains on FULL1 and FULL2 dataset, surpassing PoseNet+ by 78.3\% and 65.8\%, respectively. EffLoc exhibits impressive 33.9\% and 24.7\% improvements in FULL1 and FULL2 routes compared with MapNet by using only a single image. Additionally, when compared to AtLoc, EffLoc achieves an outstanding accuracy in all cases by a large margin. {Figure \ref{fig3}} presents the trajectory predictions of LOOP1 (top), LOOP2 (middle) and FULL1 (bottom) using Stereo VO, AtLoc, MapNet, PoseNet+, and EffLoc. Despite Stereo VO’s smooth predicted trajectories, it experiences substantial drifts as the route length extends. PoseNet+ tends to generate numerous outliers as a result of its strong reliance on local similarities. However, EffLoc significantly reduced these outliers. EffLoc guided by Sequential Group Attention (SGA) captures both local and global geometric features of the diverse and challenging real-world scenarios, enabling robust pose prediction.

\subsection{Efficiency Analysis}
To validate the efficiency of SGA mechanism in our EffLoc, we compare against AtLoc, Posenet+Mapnet, and Ms-Transformer models (Figure \ref{fig1}). EffLoc displays superior compute (Flops, Parameters) vs. accuracy trade-off. Total Flops and Parameters are Torchstat-derived. EffLoc excels with 710.95M Flops, 14.99M parameters, notably outperforming AtLoc (29.82M parameters, 5380M Flops). EffLoc-Small enhances efficiency with 11.32M parameters, 397.68M Flops. EffLoc-XS excels, with 8.66M parameters, 227.68M Flops.

EffLoc's efficiency benefits are clear, using 49.7\% fewer Flops with 9.6\% lower position and rotation error vs. AtLoc. Figure 5 illustrates convergence rate over 50 epochs, crucial for efficiency assessment. EffLoc's Sequential Group Heads refine feature representation, boosting speed 32.9\% compared to AtLoc. Figure 6 highlights EffLoc's superior total memory usage versus Atloc, Posenet+, Mapnet, MS-Transformer. Hierarchical layout with memory-bound self-attention and optimized parameters lead to EffLoc's 2.7× memory use reduction, fitting resource constraints.

In contrast to AtLoc's global feature focus, EffLoc adeptly balances local-global feature fusion, maintaining accuracy across real-world applications.

\subsection{Ablation Study}

\begin{table}
  \centering
  \footnotesize
    \begin{tabular}{c|c|c|c}
    Model & \multicolumn{1}{c}{EffLoc-XS} & \multicolumn{1}{c}{EffLoc-Small} & EffLoc \\
    \hline
    LOOP1 & 23.42m,16.82°  & 21.51m,14.67°  & \textbf{7.58m,4.12°} \\
    LOOP2 & 27.45m,20.19° & 25.39m,15.23° & \textbf{7.89m,4.19°} \\
    FULL1 & 109.23m,24.41° & 43.23m,18.77° & \textbf{27.23m,11.41°} \\
    FULL2 & 122.79m,20.47° & 57.41m,11.79° & \textbf{44.82m,9.87°} \\
    \hline
    Average & 70.72m,20.47° & 36.89m,15.12° & \textbf{21.88m,7.40°} \\
    \hline
          & \multicolumn{3}{c}{Architecture details} \\
          \hline
    $\left\{{D1, D2, D3}\right\}$ & $\left\{{128, 240, 320}\right\}$& $\left\{{128, 256, 384}\right\}$ & $\left\{{192, 288, 384}\right\}$ \\
    $\left\{{L1, L2, L3}\right\}$ & $\left\{{1, 2, 3}\right\}$ & $\left\{{1, 2, 3}\right\}$ & $\left\{{1, 3, 4}\right\}$ \\
    $\left\{{H1, H2, H3}\right\}$ & $\left\{{4, 3, 4}\right\}$ & $\left\{{4, 4, 4}\right\}$ & $\left\{{3, 3, 4}\right\}$ \\
    \end{tabular}%
    \captionsetup{font={footnotesize},justification=justified}
    \caption{Architecture details of EffLoc model variants in ablation study of EffLoc are reported. We calculate the mean errors of position and rotation and the average of different size EffLoc models. Di, Li, and Hi refer to the width, depth, and number of heads in the $i$-th stage.}
  \label{tab3}%
\end{table}%

\begin{figure}[t]
    \centering
    \includegraphics[width=\columnwidth,height=0.5\columnwidth]{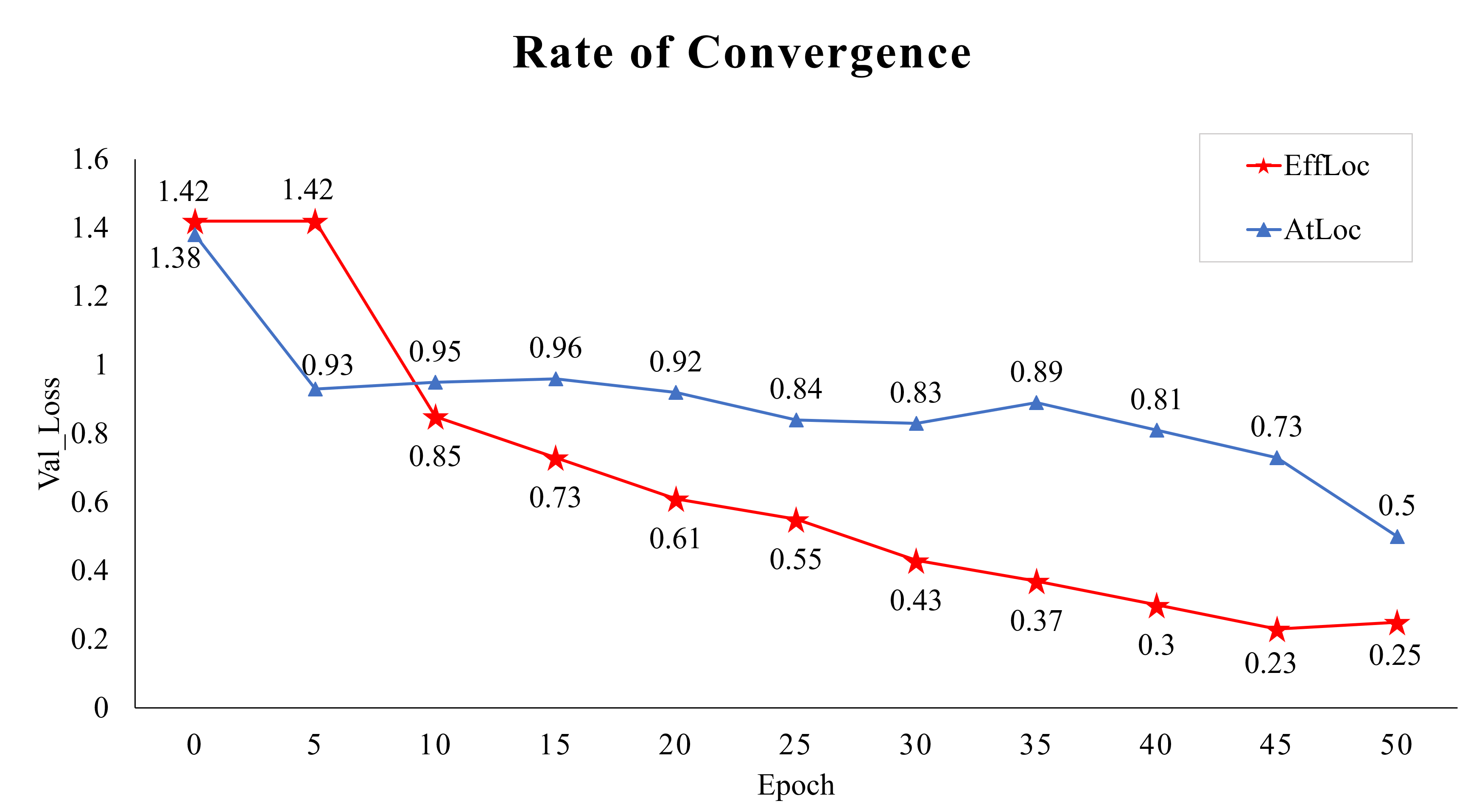}
    \captionsetup{font={footnotesize},justification=justified}
    
    \caption{Convergence velocity performances between EffLoc and AtLoc. The red line (EffLoc) of rate of convergence measures the faster speed converges to the optimal solutions as the epochs increase.}
    \label{fig4}
\end{figure}

\begin{figure}[t]
    \centering
    \includegraphics[width=\columnwidth,height=0.5\columnwidth]{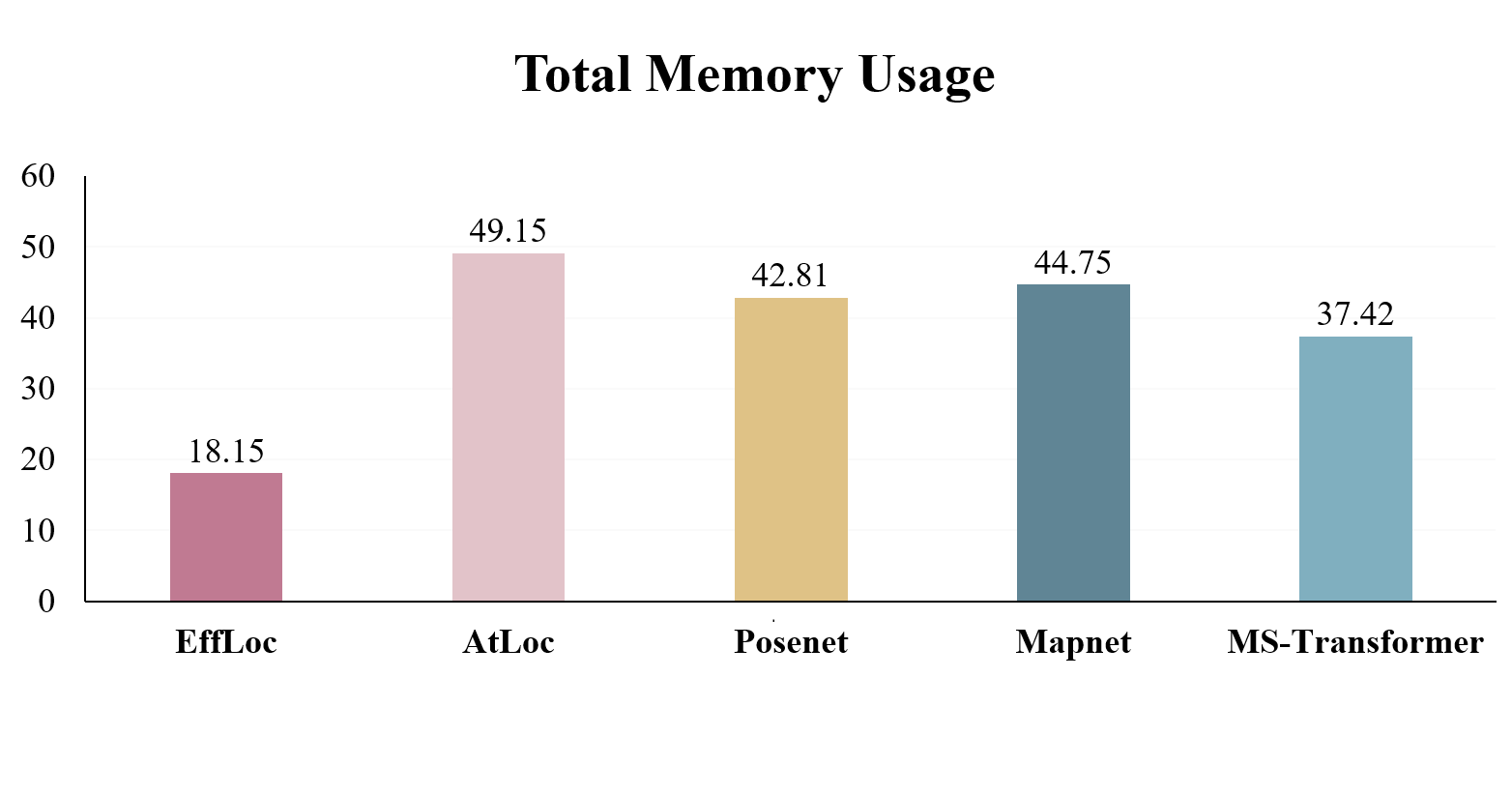}
    \captionsetup{font={footnotesize},justification=justified}
    
    \caption{ Memory Usage Comparisons with models. EffLoc optimizes memory consumption while maintaining good performance(18.15MB).}
    \label{fig5}
\end{figure}

In this section, we perform an ablation study to assess the influence of distinct architectural components in the EffLoc model using the Oxford RobotCar dataset. Architectural details are summarized in Table \ref{tab3}. We train three models with varying width, depth, and attention heads for 300 epochs, examining the accuracy-complexity trade-off. To ensure fairness, other modules remain consistent.
Table \ref{tab3} compares EffLoc, EffLoc-Small, and EffLoc-XS. EffLoc-Small has constrained channel widths (128, 240, 320) in early stages(i) and fewer blocks (1, 2, 3) initially, reducing redundancy and memory use. EffLoc-Small's hierarchy aids faster convergence but incurs slightly higher mean position (70.72m to 36.89m) and rotation errors (20.47° to 15.12°).
EffLoc's optimized parameters strike a balance between accuracy and efficiency in camera relocalization. Vital modules feature increased channels, preserving crucial feature information through higher-dimensional learning. Smaller models exhibit improved efficiency yet with marginal accuracy decline. Larger models suit unconstrained resources.
EffLoc's design aims to achieve an accuracy-efficiency tradeoff, adaptable to diverse practical scenarios. In conclusion, our ablation study emphasizes the importance of choosing appropriate width, depth, and attention heads in EffLoc for camera relocalization. Evaluation across model sizes underscores optimal accuracy-efficiency balance for real-world performance.

\section{CONCLUSIONS}
Due to the dynamic and complex nature of real-world long-distance scenes, camera relocalization poses significant challenges in the field of computer vision. Our proposed approach, EffLoc, is based on lightweight transformers and has demonstrated notable improvements in relocalization accuracy and model convergence speed, accompanied by reduced FLOPs and memory usage. In the future, we plan to focus on enhancing EffLoc's robustness and adaptability in dealing with dynamic and complex scenes.

\addtolength{\textheight}{-12cm}   




\end{document}